\documentclass{article}

\usepackage[preprint]{neurips_2025}

\usepackage[utf8]{inputenc} % allow utf-8 input
\usepackage[T1]{fontenc}    % use 8-bit T1 fonts
\usepackage{hyperref}       % hyperlinks
\usepackage{url}            % simple URL typesetting
\usepackage{booktabs}       % professional-quality tables
\usepackage{amsfonts}       % blackboard math symbols
\usepackage{nicefrac}       % compact symbols for 1/2, etc.
\usepackage{microtype}      % microtypography
\usepackage{xcolor}         % colors
\usepackage[pdftex]{graphicx}
\usepackage{subcaption}
\usepackage{float}
\usepackage{titlesec}
\titlespacing*{\section}{0pt}{0.6\baselineskip}{0.4\baselineskip} % before/after spacing

\title{Analysis of Error Sources in LLM-based Hypothesis Search for Few-Shot Rule Induction}

\author{%
  Aishni Parab\(^1,^2\) \quad
  Hongjing Lu\(^3\) \quad
  Ying Nian Wu\(^1\) \quad
  Sumit Gulwani\(^2\) \\
  \(^1\)Department of Statistics, University of California, Los Angeles \\
  \(^2\)Microsoft, Redmond, WA \\
  \(^3\)Department of Psychology, University of California, Los Angeles \\
  \texttt{\{aishni,hongjing\}@ucla.edu}, \texttt{ywu@stat.ucla.edu}, \texttt{sumitg@microsoft.com} \\
}

\begin{document}

\maketitle

\begin{abstract}
Inductive reasoning enables humans to infer abstract rules from limited examples and apply them to novel situations. In this work, we compare an LLM-based hypothesis search framework with direct program generation approaches on few-shot rule induction tasks. Our findings show that hypothesis search achieves performance comparable to humans, while direct program generation falls notably behind. An error analysis reveals key bottlenecks in hypothesis generation and suggests directions for advancing program induction methods. Overall, this paper underscores the potential of LLM-based hypothesis search for modeling inductive reasoning and the challenges in building more efficient systems.
\end{abstract}
%A recent framework by Wang et al. models this process with large language models (LLMs) that generate natural language hypotheses, implement them as programs, and refine them against input–output examples. We study this framework on the list-function induction benchmark introduced by Rule et al., which comprises 100 tasks with systematically collected human data, and compare it with direct program generation.
\section{Introduction}
\label{intro}

Humans can infer general rules from only a handful of examples and flexibly apply them to novel situations, a process known as inductive reasoning \citep{peirce1868questions}. This human ability in few-shot rule induction is striking because  many different rules could explain the same limited observations, yet people tend to converge on explanations in a systematic way. Rule induction can be understood as a search through a hypothesis space: generating candidate hypotheses about rules, testing them against observations, and updating their beliefs as evidence accumulates \citep{restle1961psychology}. The efficiency of this process depends both on how the hypothesis space is represented and on how learners search within it \citep{griffiths2010probabilistic, tenenbaum2011grow}. Programs provide a natural representation of hypotheses, and searching the program space implements the process of considering alternative explanations, and execution supplies a direct test against data \citep{tenenbaum2011grow}. 

A wide range of methods, including Bayesian approaches \citep{tenenbaum2006theory, lewis2014error}, symbolic search \citep{gulwani2011automating,feser2015synthesizing}, end-to-end neural approaches \citep{parisotto2016neuro, devlin2017robustfill, kalyan2018neural,valkov2018houdini}, and neuro-symbolic hybrids \citep{ mao2019neuro, ellis2021dreamcoder}, have been developed to implement rule induction as program generation. More recently, Large Language Models (LLMs) have emerged as powerful engines for program generation \citep{bubeck2023sparks,  austin2021program, poesia2022synchromesh}. Recent work has leveraged LLMs' ability of in-context learning with few examples \citep{brown2020language, li2024programming}, translate natural language into code \citep{chen2021evaluating}, refine candidate programs through iterative self-improvement \citep{shinn2024reflexion}, and coordinate specialized modules in multi-agent architectures \citep{talebirad2023multi, webb2023prefrontal, piriyakulkij2024doing, grayeli2024symbolic}. \citet{wang2023hypothesis} integrated LLMs into a hypothesis-driven search pipeline for the Abstraction and Reasoning Corpus (ARC), a benchmark that evaluates inductive reasoning through abstract two-dimensional grid transformations \citep{chollet2019measure}. Building on Bayesian views of inductive reasoning \citep{tenenbaum2011grow, goodman2011learning}, they used LLMs to generate rule-like hypotheses in natural language, translate them into candidate programs, and iteratively refine them against observed input–output examples. Each candidate program was validated by execution, and passing programs were retained for testing. The authors found that this LLM-based hypothesis search framework performs better in few-shot concept induction than direct program generation approaches \citep{mirchandani2023large, gendron2023large, xu2023llms, johnson2021fast}. 

ARC has become a central benchmark for testing inductive reasoning ability in AI systems, yet its grid-based format introduces perceptual complexity that can obscure the core challenge of rule induction. To focus directly on this challenge, we examine a simpler domain of list-function induction, introduced by \citet{rule2024symbolic}, where the task is to infer the transformation rule from input–output lists. This domain presents systematically measured human performance on 100 functions, enabling direct comparison with models of rule induction and human learners.

Our contributions are threefold. First, we evaluate the hypothesis search framework of \citet{wang2023hypothesis} on the 100 list-function tasks, aligning model evaluation with the human benchmark. Second, we compare hypothesis search to direct program generation, showing it performs better and achieves human-comparable acquisition performance. Third, we present a detailed error analysis of the hypothesis search pipeline, identifying bottlenecks in generation, summarization, and refinement. Together, these results provide a systematic case study that highlights both the narrow conditions under which large language models approximate human reasoning and the deeper reasons that currently prevent them from serving as reliable models of inductive reasoning.

%\section{Related Work - Incomplete}
%\label{relatedwork}
%	•	Inductive reasoning as program synthesis (cite program induction, ARC, meta-primitive).
%	•	Hypothesis search paradigm (Goodman et al.).
%	•	Human benchmarks (Nature paper on meta-primitives).
%	•	Other    Human inductive reasoning and program induction (summarize psychology/cogsci foundations, Bayesian models).
%	•	ARC benchmark and beyond (situate your contribution relative to ARC and other reasoning tasks).
%	•	LLM-based program induction (review prior work, including direct generation, search-based, and hybrid methods).

\begin{figure}[ht]
  \centering
  \begin{minipage}{0.78\linewidth}
    \centering
    \includegraphics[width=\linewidth]{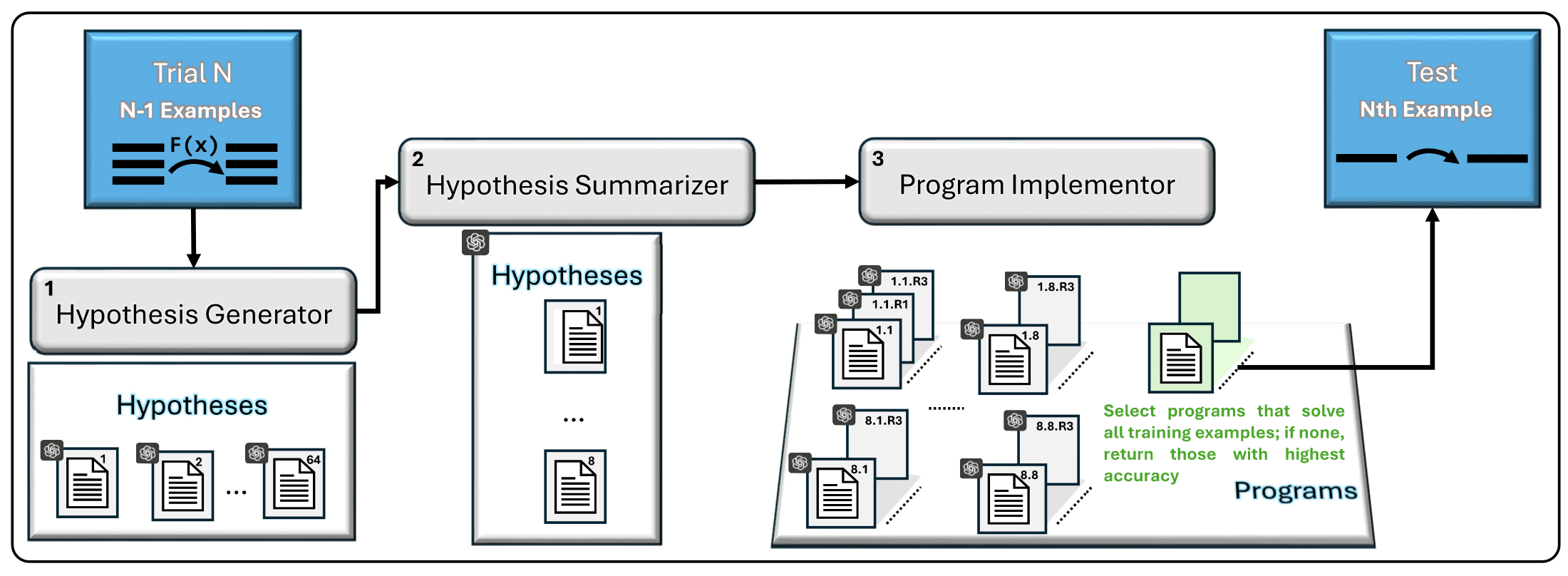}
    \caption{Hypothesis Search Framework proposed by Wang et al \citet{wang2023hypothesis} adapted to experiment protocol of \citet{rule2024symbolic}.}
    \label{fig:pipeline}
  \end{minipage}\hfill
  \begin{minipage}{0.21\linewidth}
    \centering
    \includegraphics[ width=\linewidth, height=3.2cm]{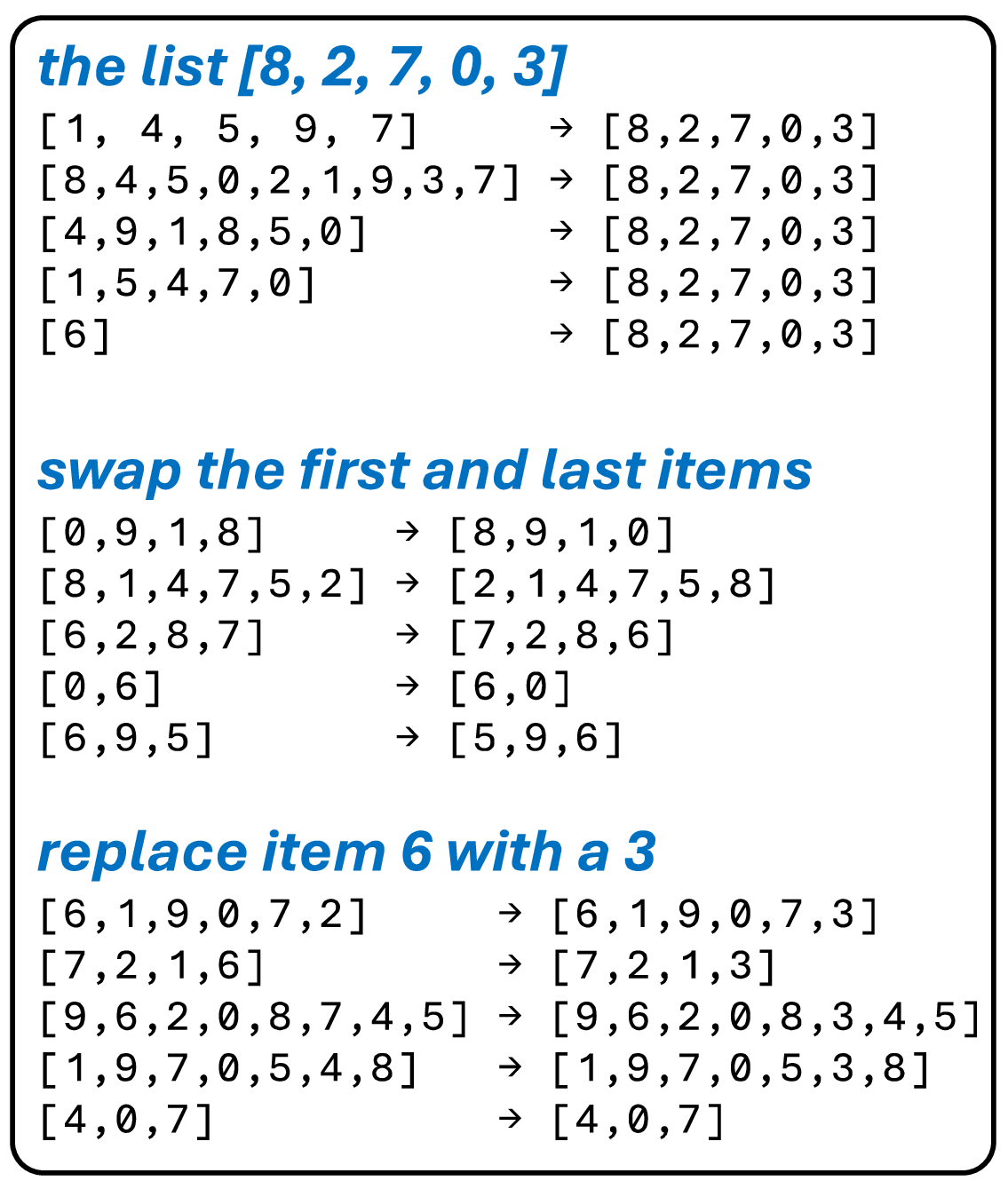}
    \caption{Three examples of list transformations in \citet{rule2024symbolic}}
    \label{fig:examples}
  \end{minipage}
\end{figure}

\section{Methods}
\label{method}
\subsection{List Function Induction}
We evaluate models using the few-shot rule induction task of list-functions, where each task specifies a mapping from input lists of integers to corresponding output lists, governed by a transformation rule. We use the 100 list functions of \citet{rule2024symbolic} (Figure~\ref{fig:examples}), which also report human performance. Each task has 11 trials: $n-1$ training examples precede prediction on trial $n$.

We evaluate both hypothesis search and direct program generation using LLMs on these 100 list-function tasks, following the evaluation protocol of \citet{rule2024symbolic}. We repeat the simulation experiment for five runs and report the mean number of functions acquired at each trial, averaged across runs (Supplementary B)

\subsection{Hypothesis Search}
We implement the hypothesis search framework introduced by \citet{wang2023hypothesis}, illustrated in Figure~\ref{fig:pipeline}. The framework consists of three LLM modules: a Hypothesis Generator, a Hypothesis Summarizer, and a Program Implementor. 

Given $n$ input-output examples of a list transformation, the \textit{Generator} produces a natural language hypothesis of how to make the list transformation. The prompt (Supplementary Figure A.2) provides demonstrations of example–hypothesis pairs. For each task, the Generator is queried 64 times, yielding 64 independent hypotheses.

The \textit{Summarizer} then condenses these hypotheses (Supplementary Figure A.3). It receives the list of 64 hypotheses and queries the model to produce 8 distinct summaries, each representing a candidate transformation rule.

For each summarized hypothesis, the \textit{Program Implementor} generates and refines executable programs following the procedure of Algorithm 1 in \citet{wang2023hypothesis} (Supplementary Figure A.4). Initially, it produces 8 candidate Python programs per hypothesis. Each program is evaluated against the provided input–output examples. If a program passes all examples, it is selected for testing. If not, the Implementor enters a refinement phase.

In the refinement phase, each candidate program may be revised for up to three rounds. At every round, the program is executed on all training examples. When an evaluation error (mismatch between predicted and target outputs) or an execution error (syntax or runtime failure) occurs, the error message is appended to the prompt query, and the model generates a refined program (Supplementary Figure A.5). If a refined program solves all examples, it is returned as the solution. Otherwise, refinement continues until a program either succeeds or the round limit is reached. 

The Implementor iterates through all 8 candidate hypotheses, maintaining the program that achieves the highest training accuracy. If no perfect solution emerges, the best-performing program is selected. The final program is then tested on the held-out test example, and task accuracy is reported. When several programs achieve the same maximum training accuracy, we modify the original algorithm by testing all tied programs and reporting their average test accuracy, rather than returning only the first best.

\subsection{Direct Program Generation}
Direct program generation models directly generate executable code from given examples rather than passing through a hypothesis search pipeline. Given $n$ input-output examples, the model is prompted to produce a Python program that implements the transformation consistent with those examples (Supplementary Figure A.1). The generated program is executed on both the training examples and the held-out test example, and its accuracy is reported. We used GPT-4o and Codex as the baseline model of direction program generation approaches. 

\section{Results}

\begin{figure}[ht]
  \centering
  % Left figure
  \begin{subfigure}[t]{0.48\linewidth}
    \centering
    \includegraphics[width=\linewidth]{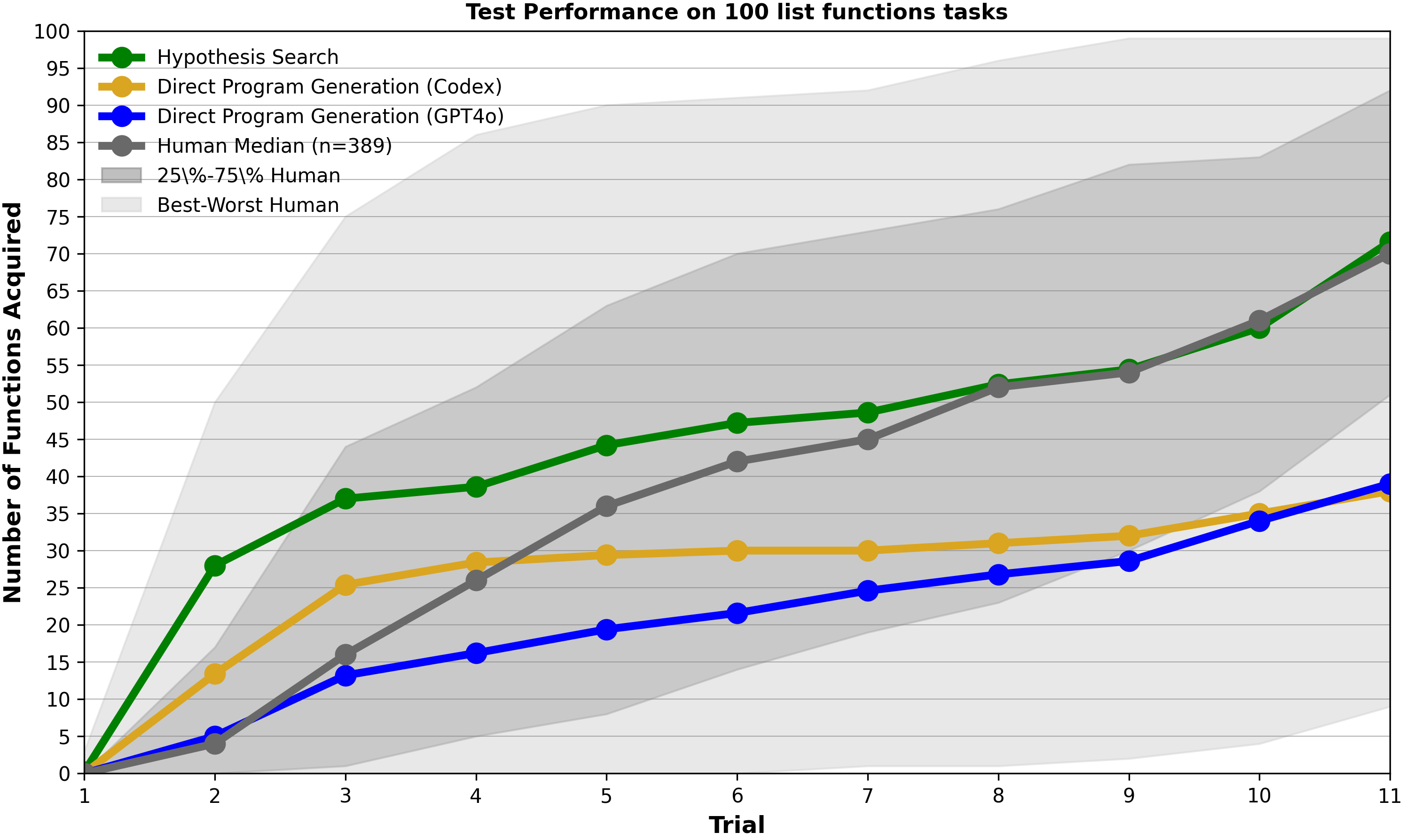}
    \caption{Acquisition curves of test accuracy across 11 trials. 
    Humans and hypothesis search display similar learning performance, 
    while direct program generation methods lag behind. Hypothesis Search acquires more functions with fewer observations.}
    \label{fig:acquisition}
  \end{subfigure}
  \hfill
  % Right figure
  \begin{subfigure}[t]{0.48\linewidth}
    \centering
    \includegraphics[width=\linewidth]{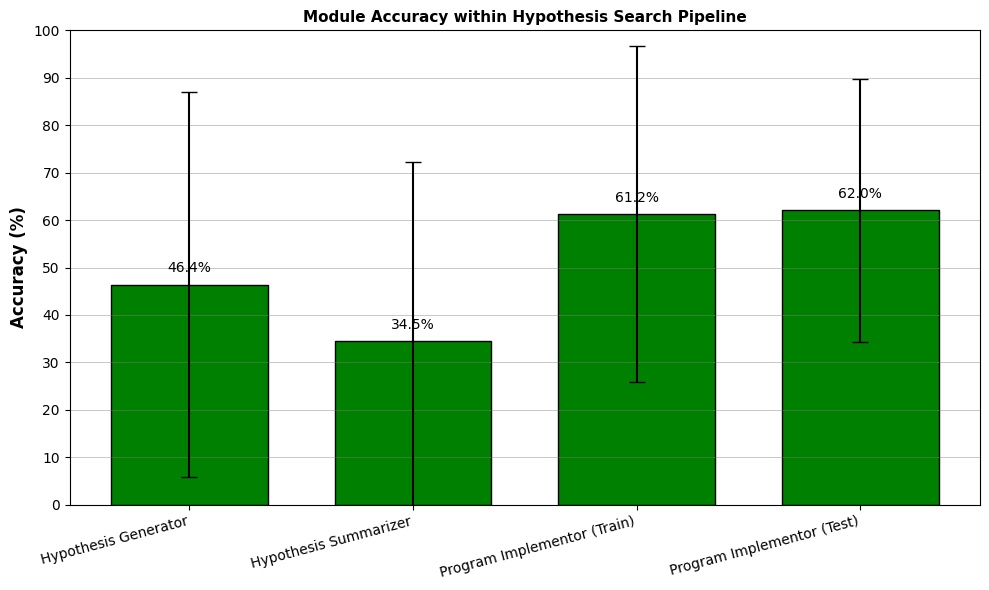}
    \caption{Module accuracy within the hypothesis search pipeline. 
Bars show mean accuracy with standard deviation. 
Each module is evaluated according to the success criteria described in Section~\ref{error-analysis}.}
    \label{fig:module_acc}
  \end{subfigure}
  \caption{Comparison of acquisition performance (left) and module-level accuracy (right).}
  \label{fig:acquisition_and_modules}
\end{figure}

Figure~\ref{fig:acquisition} presents acquisition curves averaged over five runs for each trial for models and humans. The $y$-axis reports the number of functions acquired out of 100 as the number of training examples, i.e., trial, increases. The acquisition performance is defined as the number of tasks successfully solved at least once in a given trial. Humans and hypothesis search follow similar trajectories, with the number of acquired functions increasing steadily as more examples are observed. Direct program generation performs worse, with GPT-4o and Codex solving fewer tasks overall and showing moderate improvement as additional examples are provided. These findings show that hypothesis-guided search outperforms direct program generation, highlighting the benefit of using natural language hypotheses to guide the search over candidate programs for few-shot rule induction (see Supplementary Table 1 for aggregate results).

While hypothesis search achieves human-comparable performance, the computation cost is high. Each task requires 64 LLM calls in the Hypothesis Generator, one call in the Hypothesis Summarizer, and up to $8 \times 8 \times 3$ calls in the Program Implementor for candidate generation and refinement, resulting in as many as $257$ calls per task in the worst case. These costs highlight the importance of understanding where errors arise in the Hypothesis Search pipeline. The next section analyzes the Generator, Summarizer, and Implementor modules to identify bottlenecks and guide improvements in both efficiency and reasoning performance.

\section{Error Analysis of the Hypothesis Search Pipeline}
\label{error-analysis}
To locate sources of error in the hypothesis search pipeline, we evaluate each module in isolation. For the Hypothesis Generator and Summarizer, we use an additional LLM module as an Evaluator (Supplementary Figure A.6). \citet{rule2024symbolic} provide a natural language description of each task, which we treat as the ground truth. The Generator produces 64 hypotheses per trial, which the Summarizer condenses to 8. Each of these hypotheses is scored by the Evaluator as correct or incorrect based on its match to the ground-truth description. A trial is scored as successful if at least one correct hypothesis is identified in the modules. 

%The Program Implementor receives up to 8 summarized hypotheses and produces 8 initial candidate programs for each. If none of these solve all training examples, each candidate can undergo up to three rounds of refinement. 
We evaluate the Program Implementor on training data, where a trial is successful if at least one candidate program solves all examples. We then repeat the evaluation on held-out test data; details are provided in Appendix C–E, and the findings are consistent.

Supplementary Table 2 reports average accuracies for each module, and Figure~\ref{fig:module_acc} shows the module accuracy. The Hypothesis Generator achieves 46.4\% accuracy, 
%(add up gen success rows indices 9 - 16 then divide by 5500)
while the Summarizer drops to 34.5\%. 
%(add up rows 5-8 and 13-16 then divide by 5500)
The Program Implementor achieves 61.2\% accuracy, demonstrating the strength of program synthesis and refinement even when upstream hypotheses are imperfect. Notably, the Summarizer discards 26\% of correct generator hypotheses, retaining only 74\%, which accounts for an 12.1\% absolute loss in overall trial performance. 
%(rows 9-12 divided by 5500)

As reported in Supplementary Table 3, the overall \emph{test} solve rate is 62.0\% and the failure rate is 38.1\%. Several patterns emerge. \textit{First}, the pipeline depends on the Generator: there are no trials in which the Summarizer succeeds while the Generator fails, confirming that Generator accuracy is foundational. \textit{Second}, when the Generator succeeds but the Summarizer fails, the Implementor still produces a correct \emph{test} program in 75.3\% cases. %((52+449)/(115+52+49+449)\). 
\textit{Third}, even under double failure at the front end, the Implementor rescues 1092 trials on the test example, which is 19.9\% of all trials %\((1092/5500)\) 
and 37.0\% of the double-failure subset. %\((1092/2948)\) 
Finally, when both the Generator and Summarizer succeed, the Implementor solves the test example in 95.9\% of cases. %\((14+1804)/(42+14+36+1804)\).
Overall, the Implementor provides meaningful robustness to upstream errors; Supplementary D details its refinement phase and computational trade-offs.

Nevertheless, Generator accuracy strongly predicts downstream success. At the trial level, Generator and Implementor outcomes correlate at $r=0.57$ on the training set and $r=0.55$ on the test set. Aggregated by task, these correlations rise to $r=0.93$ and $r=0.79$, respectively, underscoring the close link between hypothesis quality and Implementor performance. Logistic regression further quantifies this dependency. Implementor success on the training set is about 19 times more likely when the generator produces a correct hypothesis ($p < 0.001$), while test success is about 17 times more likely under the same condition ($p < 0.001$). The slight drop from train to test highlights cases where brittle training fits fail to generalize, but across both measures the effect is substantial. Together, these analyses reinforce the Generator as the central bottleneck in the pipeline and show that improvements at this stage would yield the largest downstream gains.

\section{Conclusion}
We presented a systematic analysis of hypothesis search for inductive list-function learning, comparing human performance, direct program generation, and LLM-based hypothesis search. Our results show that hypothesis search yields human-comparable acquisition performance but remains limited by bottlenecks in hypothesis generation and by the cost of program refinement. Future work should focus on improving the quality of generated hypotheses and on developing more efficient program search strategies, with the broader goal of building reasoning systems that combine human-like flexibility with scalable computational efficiency.

\bibliography{references}
\bibliographystyle{plainnat}

\clearpage
\appendix
\section*{Supplementary Material}
\addcontentsline{toc}{section}{Supplementary Material}

\section{Model Prompts}
\label{sec:prompts}

\begin{figure}[htbp]
  \centering
\includegraphics[width=\linewidth]{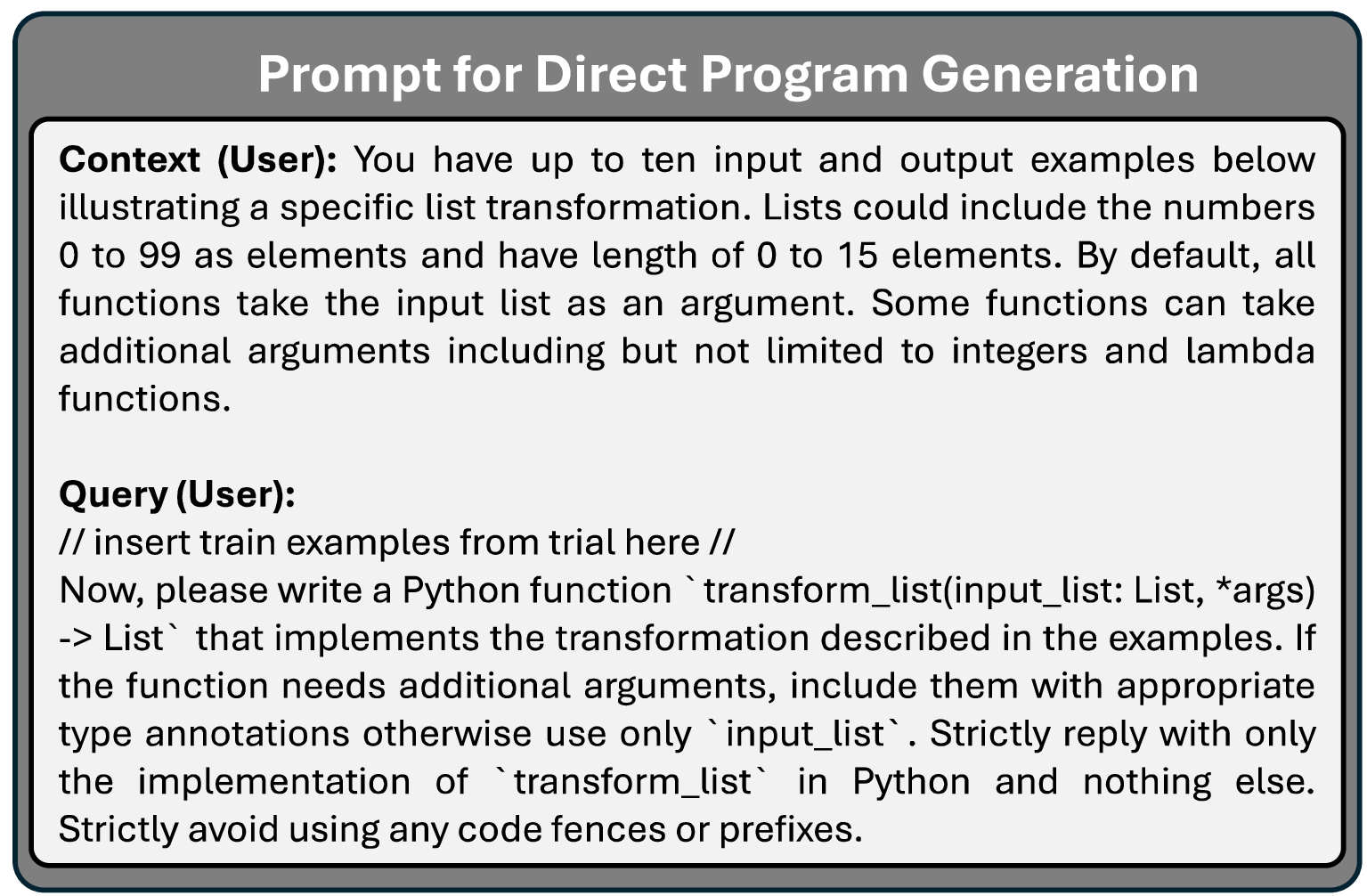}
  \caption{This prompt is used to directly generate a program given input-output examples.}
  \label{fig:direct_prompt}
\end{figure}

\begin{figure}[htbp]
  \centering
\includegraphics[width=\linewidth]{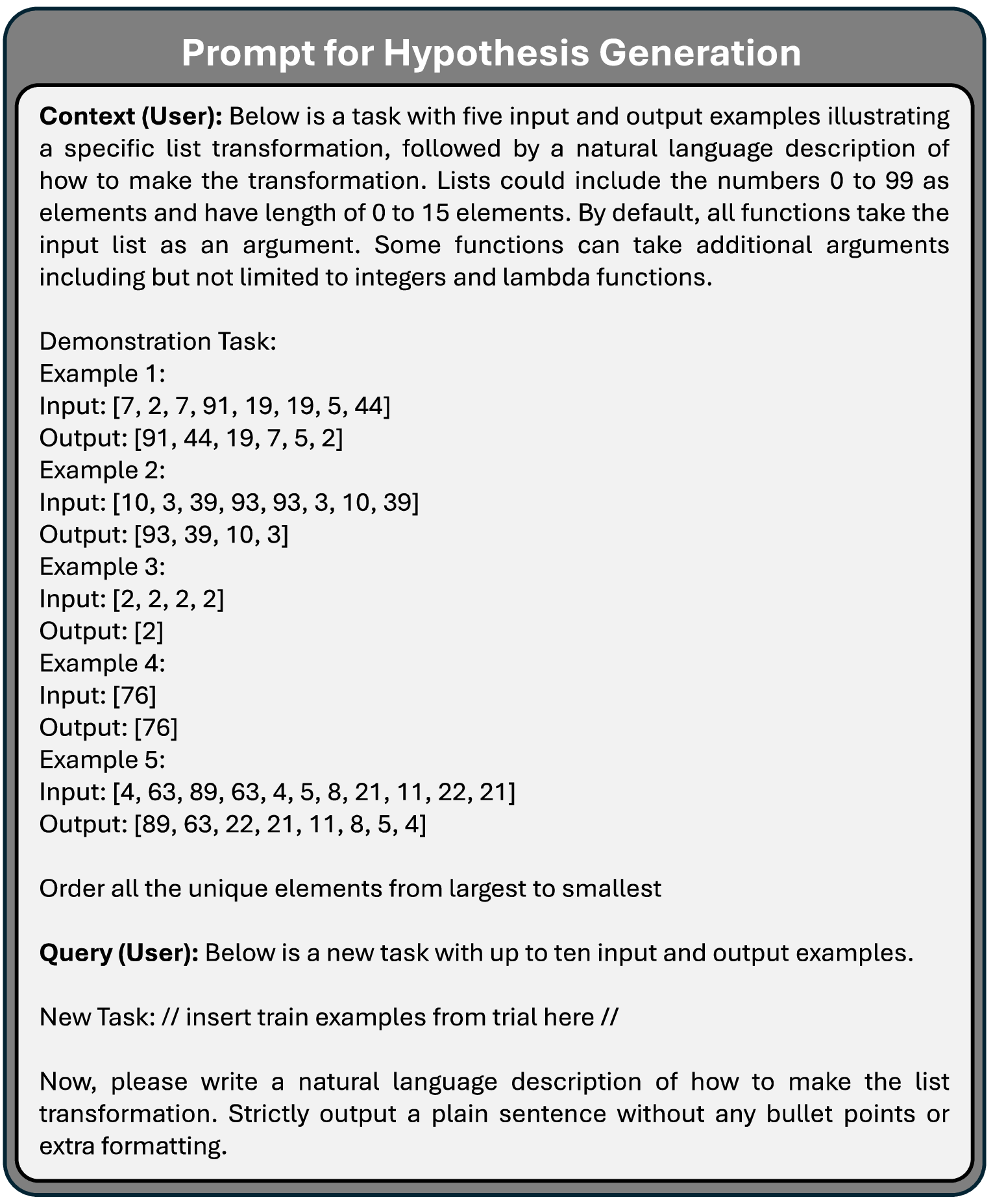}
  \caption{The prompt used by the \textit{Generator} module in the Hypothesis Search pipeline.}
  \label{fig:generator_prompt}
\end{figure}

\begin{figure}[htbp]
  \centering
\includegraphics[width=\linewidth]{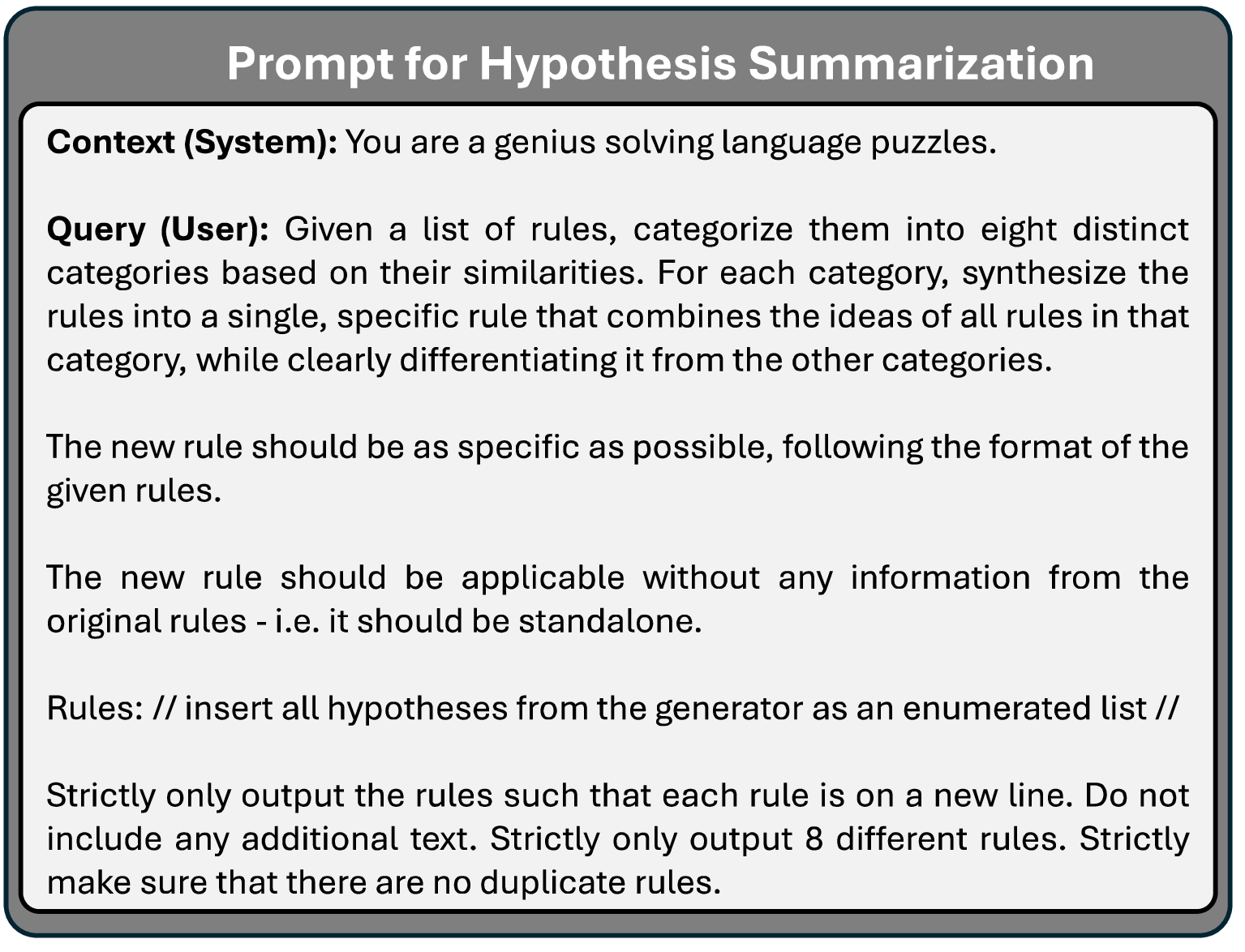}
  \caption{The prompt used by the \textit{Summarizer} module in the Hypothesis Search pipeline.}
  \label{fig:summarizer_prompt}
\end{figure}

\begin{figure}[htbp]
  \centering
\includegraphics[width=\linewidth]{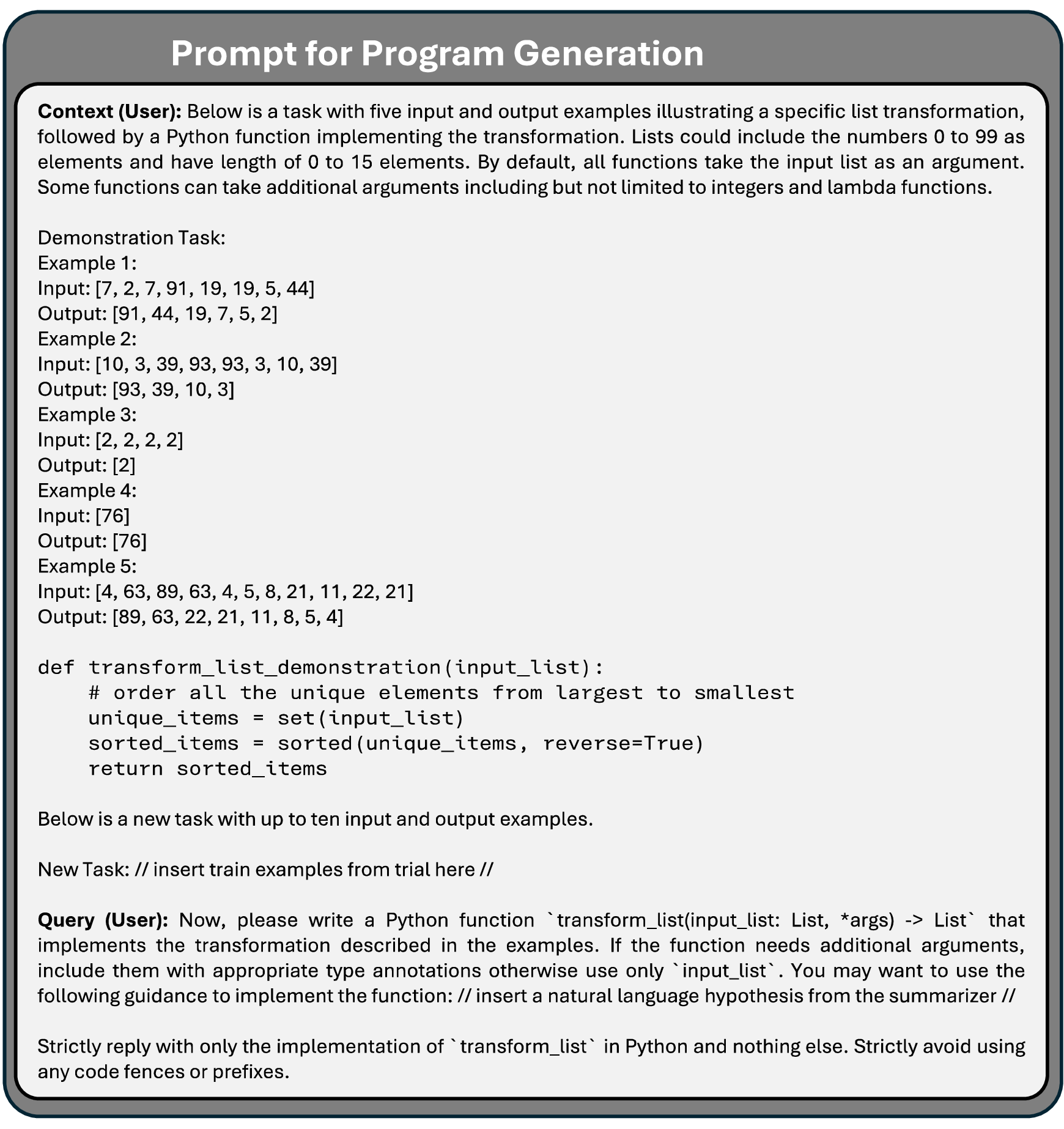}
  \caption{The prompt used by the \textit{Program Implementor} in the Hypothesis Search pipeline module to generate initial program candidates.}
  \label{fig:program_prompt}
\end{figure}

\begin{figure}[htbp]
  \centering
\includegraphics[width=\linewidth]{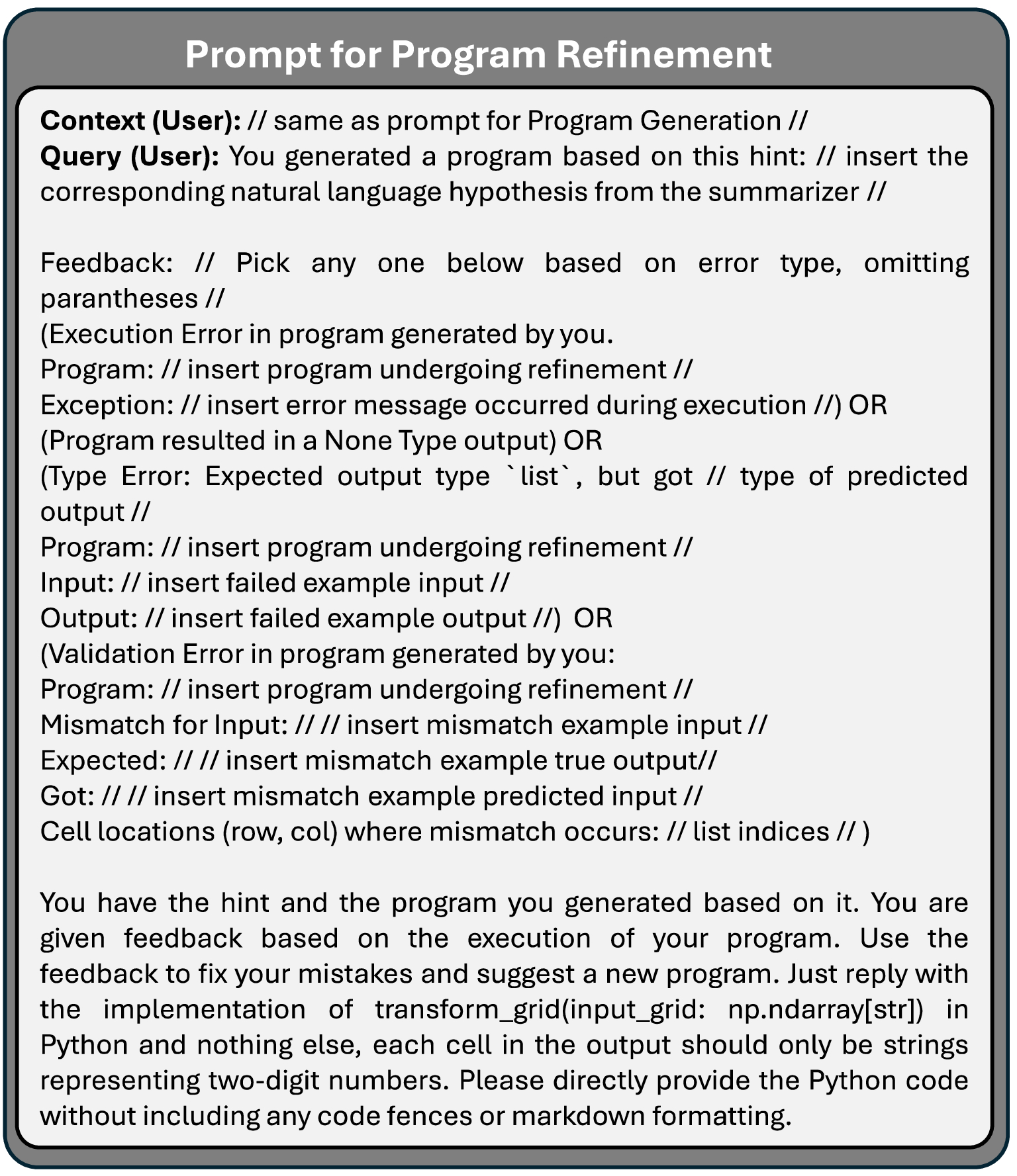}
  \caption{Query Prompt used by the \textit{Program Implementor} in the Hypothesis Search pipeline for refinement.}
  \label{fig:refinement_prompt}
\end{figure}

\begin{figure}[htbp]
  \centering
\includegraphics[width=\linewidth]{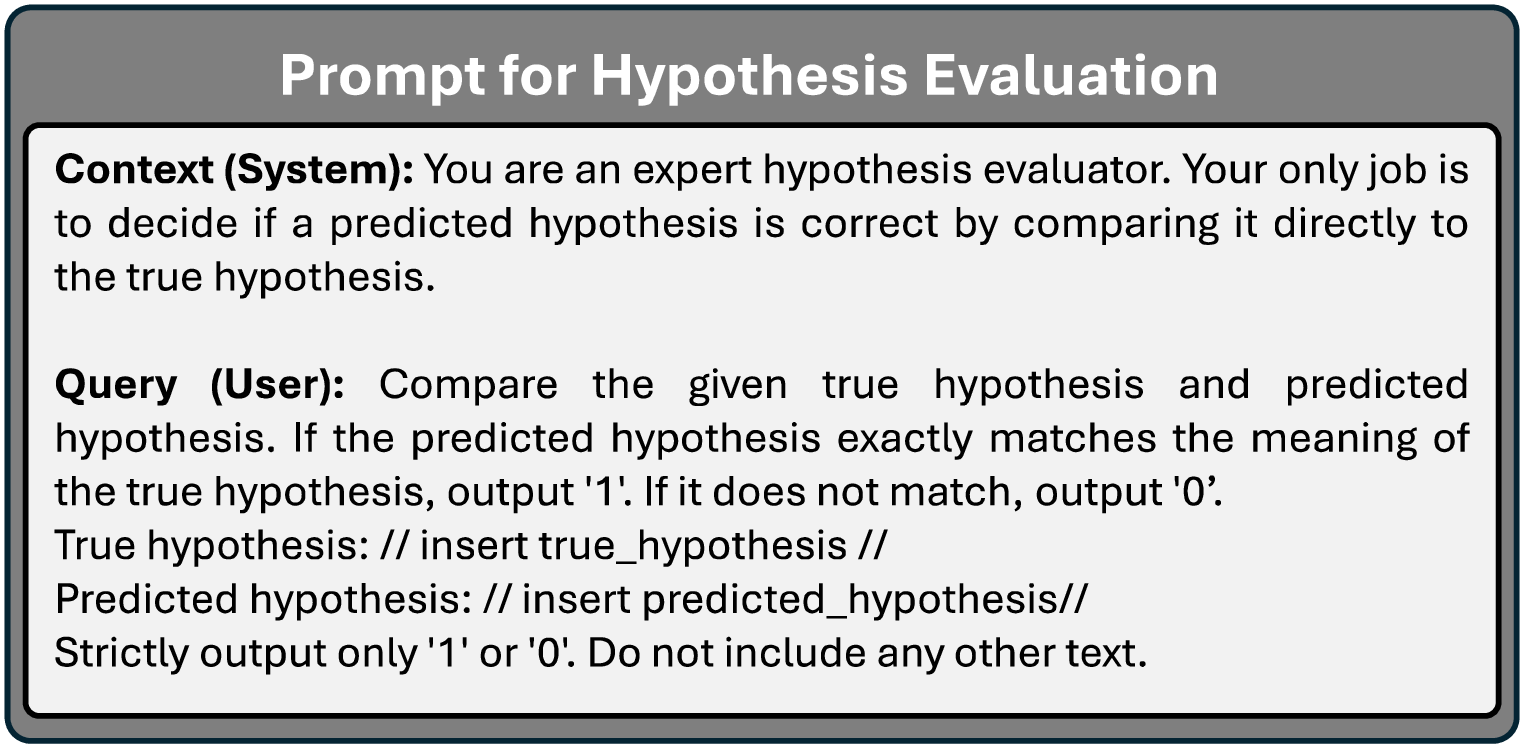}
  \caption{Prompt used to evaluate hypotheses resulting from the \textit{Generator} and \textit{Simulator} modules of the Hypothesis Search pipeline.}
  \label{fig:refinement_prompt}
\end{figure}

\newpage
\section{GPT simulation parameters}
\label{gpt_params}
Our experiments consider two GPT-4o conditions. In Direct Program Generation, GPT-4o is run with temperature 0. In Hypothesis Search, we follow the configuration of \citet{wang2023hypothesis}, using GPT-4o for all three modules: the Hypothesis Generator (temperature 1.0, top-p 1.0), the Hypothesis Summarizer (temperature 1.0, top-p 0.0), and the Program Implementor (temperature 0.7, top-p 0.0). All modules use a maximum token limit of 1000. For comparison, we also report human performance and Codex results from \citet{rule2024symbolic}, where Codex was evaluated under Direct Program Generation.

While \citet{rule2024symbolic} represent list functions in a functional language over abstract primitives, we instead generate candidate programs directly in Python and evaluate them against the provided examples in all experiments.

\clearpage
\section{Results: Mean Test Accuracy}
\label{test_results}
\begin{table}[t]
  \caption{Mean test accuracy on the 100 list-function tasks. Results for humans and Codex are reported from \citet{rule2024symbolic}}
  \label{tab:overall-results}
  \centering
  \begin{tabular}{lcc}
    \toprule
    Model & Mean Test Accuracy & Std \\
    \midrule
    Human \citep{rule2024symbolic} & 0.521 & 0.202 \\
    Hypothesis Search & 0.487 & 0.326 \\
    Direct Gen (GPT4o) & 0.359 & 0.309 \\
    Direct Gen (Codex) \citep{rule2024symbolic} & 0.322 & 0.467 \\
    \bottomrule
  \end{tabular}
\end{table} 

Table~\ref{tab:overall-results} reports mean test accuracy, which measures per-example correctness averaged across tasks. Human learners reach 0.521, slightly higher than hypothesis search at 0.487. Direct program generation performs substantially worse, with GPT-4o reaching 0.359 and Codex 0.322. Standard deviations further show that human and hypothesis search results are more consistent across tasks than direct generation. 

In the Results, the two reported metrics, mean model performance and function acquisition per trial, highlight a tradeoff. While acquisition curves capture breadth, \textit{how many tasks the system covers}, mean accuracy captures depth, \textit{how fully solutions generalize once acquired}. Hypothesis search achieves broader coverage than humans but often relies on brittle programs that lower average accuracy. Humans acquire fewer tasks in aggregate but typically achieve higher accuracy once they succeed. 

\begin{figure}[H]
  \centering
  %\fbox{\rule[-.5cm]{0cm}{4cm} \rule[-.5cm]{4cm}{0cm}}
 \includegraphics[width=\linewidth]{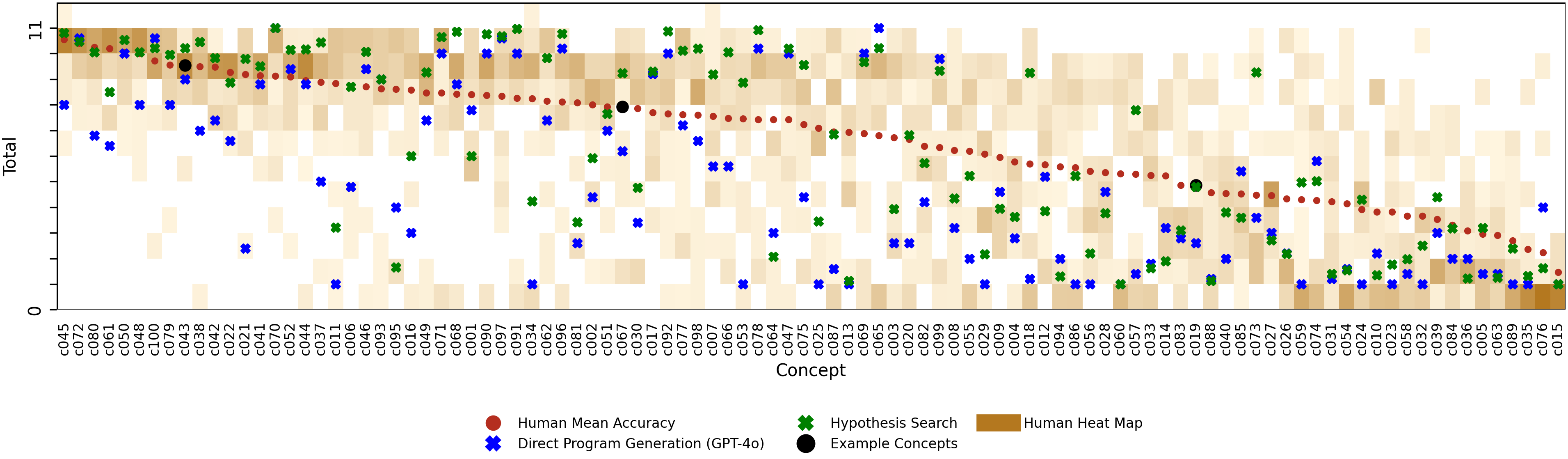}
  \caption{Task-level performance. Human accuracy varies widely across functions. Hypothesis search aligns with human difficulty patterns on many tasks but diverges on a subset. Direct program generation shows larger gaps, failing on tasks that humans find easy.}
  \label{fig:difficulty}
\end{figure}

Figure~\ref{fig:difficulty} examines performance at the level of individual tasks. Human accuracy varies widely across functions, with some acquired almost universally and others remaining extremely challenging. Hypothesis search captures part of this distribution, aligning with human performance on many tasks (e.g. c100, c071, c017, c019, c083),  but failing sharply on a subset (e.g. c045, c080, c021, c053, c018). Direct program generation models show even larger discrepancies, often failing on tasks that humans find straightforward (e.g. c072, c079, c044, c051, c078). This divergence highlights systematic weaknesses in direct generation.

\clearpage
\section{Results: Errors Associated with Program Implementor}
\label{implementor_errors}
\begin{table}[t]
  \caption{Accuracy of each module in the hypothesis search pipeline.}
  \label{tab:module-accuracy}
  \centering
  \begin{tabular}{lcc}
    \toprule
    Module & Mean Accuracy & Std \\
    \midrule
    Hypothesis Generator & 0.464  & 0.406 \\
    Hypothesis Summarizer & 0.345 & 0.377 \\
    Program Implementor (Train) & 0.612 & 0.355 \\
    Program Implementor (Test) & 0.620 & 0.278 \\
    \bottomrule
  \end{tabular}
\end{table}

We evaluate the Program Implementor in two complementary ways. First, at the \textit{training level} reported in the main text, a trial is considered successful at training time if at least one candidate program solves all training examples. This measures the module’s ability to search the space of programs consistent with observed data. All programs that achieve highest training accuracy are selected for testing on one held-out example. If atleast one of these programs solve the test example, we consider the trial successful at test time. This captures the module’s contribution to end-to-end generalization. The accuracy of the program implementor module is similar for training and test data, as shown in table below (see Table~\ref{tab:module-accuracy}). Table~\ref{tab:error-breakdown2} shows the error patten for both metrics of the Program Implementor module. 

\begin{figure}[t]
  \centering
  \includegraphics[width=0.6\linewidth]{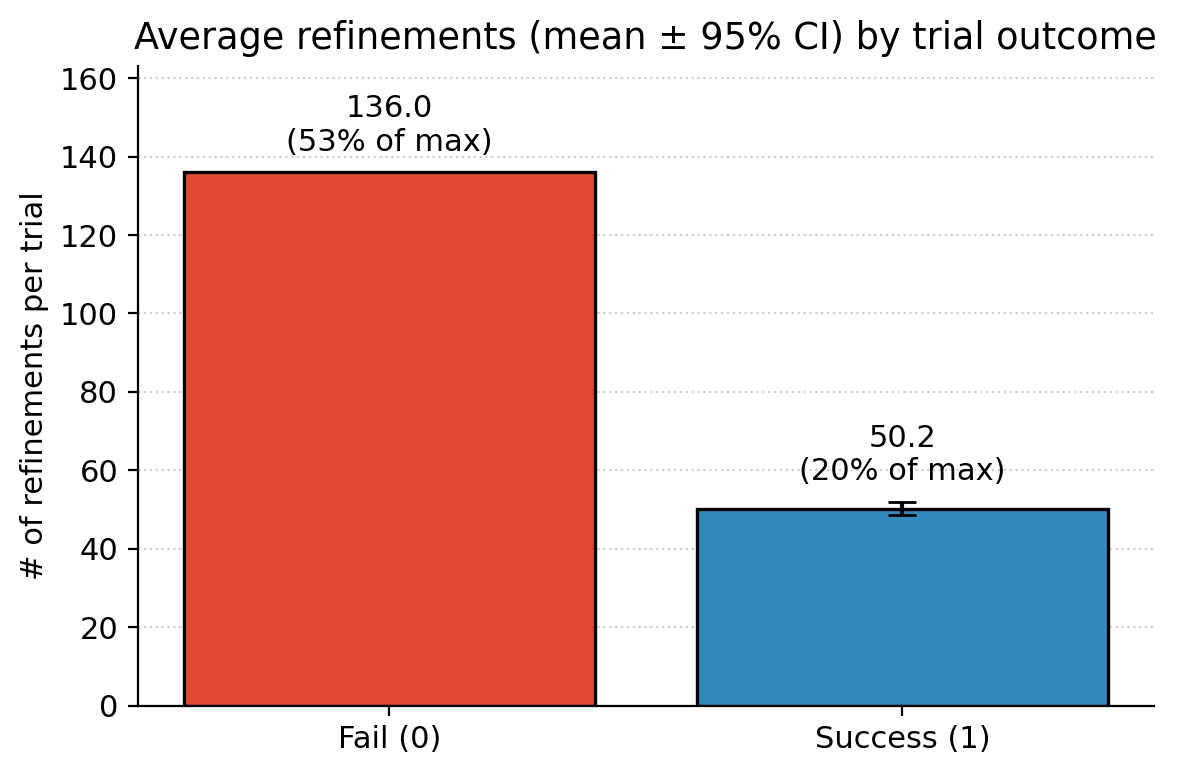}
  \caption{Average number of refinements per trial, separated by outcome. 
  Failed trials require more than twice as many refinements as successful ones, 
  highlighting the disproportionate computational cost of failure.}
  \label{fig:refinements_avg}
\end{figure}

Each trial admits up to 256 program refinements (8 hypotheses × 8 candidates × 4 versions each). When measured on training examples, successful trials required only about 50 refinements on average (roughly 20\% of the budget), whereas failed trials consumed 136 (about 53\%) (see Figure~\ref{fig:refinements_avg}). When measured on test outcomes, the asymmetry remains but is less pronounced: successful trials averaged 76 refinements (about 30\%), while failed ones consumed 95 (about 37\%). This attenuation reflects the evaluation setup. At test time, only the best-performing programs from training are selected, and a trial is counted as successful if at least one of them solve the single held-out example. This makes test success easier to achieve than training success, since solving one example does not guarantee robust generalization. Across both metrics, however, failed trials consistently require more refinement effort. This finding underscores the inefficiency of relying on downstream search to compensate for upstream errors.

\begin{table}[t]
  \caption{Error analysis across modules. Each row indicates success or failure of the four modules and the corresponding count and proportion.}
  \label{tab:error-breakdown2}
  \centering
  \begin{tabular}{lccccccc}
    \toprule
    Index & Generator & Summarizer & Implementor (Train) & Implementor (Test) & Count & Proportion (\%) \\
    \midrule
    1  & Failure & Failure & Failure & Failure & 1116 & 20.3 \\ 
    2  & Failure & Failure & Failure & Success & 793   & 14.4  \\ 
    3  & Failure & Failure & Success & Failure & 740  & 13.2 \\ 
    4  & Failure & Failure & Success & Success & 299  & 5.4  \\ 
    5  & Failure & Success & Failure & Failure & 0    & 0.0  \\ 
    6  & Failure & Success & Failure & Success & 0    & 0.0  \\ 
    7  & Failure & Success & Success & Failure & 0    & 0.0  \\ 
    8  & Failure & Success & Success & Success & 0    & 0.0  \\ 
    9  & Success & Failure & Failure & Failure & 115  & 2.1  \\ 
    10 & Success & Failure & Failure & Success & 52    & 0.9  \\ 
    11 & Success & Failure & Success & Failure & 49  & 0.7  \\ 
    12 & Success & Failure & Success & Success & 449  & 8.2  \\ 
    13 & Success & Success & Failure & Failure & 42   & 0.8  \\ 
    14 & Success & Success & Failure & Success & 14    & 0.3  \\ 
    15 & Success & Success & Success & Failure & 36  & 0.7  \\ 
    16 & Success & Success & Success & Success & 1804 & 32.8 \\ 
    \midrule
    \textbf{Total} &  &  &  &  & 5500 & 100.0 \\
    \bottomrule
  \end{tabular}
\end{table}

Refinement rarely rescued semantically incorrect hypotheses: only 4.3\% of such cases yielded a program that solved all training examples. In other words, refinement occasionally found a program consistent with the observed data, but this reflects success at training-time search rather than recovery of the true underlying rule. Because programs are evaluated on only a single held-out test example, we cannot directly measure whether such rescues translate into genuine generalization. This pattern suggests that refinement primarily corrects execution-level errors (e.g., syntax or runtime issues) rather than semantic mistakes in the hypothesized transformation rule.

While refinement recovers some upstream errors, Generator accuracy ultimately determines downstream success. The summarization stage never produces a correct output without a correct Generator hypothesis, so increasing Generator accuracy above its current 46.4\% would directly enlarge the set of near-guaranteed successes. The Implementor contributes robustness by rescuing more than one third of double-failure cases, yet nearly two thirds remain unresolved and up to 257 calls per trial may be required, making heavy reliance on refinement computationally costly.

Despite these limitations, Hypothesis Search achieves significantly better performance in few-shot rule induction than Direct Program Generation. This advantage arises because the Generator and summarization modules constrain the hypothesis space, and the Implementor samples and refines programs within it; by restricting the set of candidate programs, natural language hypotheses increase the likelihood of identifying the correct transformation rule, even when the hypotheses are imperfect. However, interpreting these gains requires caution, as the evaluation protocol introduces its own challenges.

In the current setup \citet{rule2024symbolic}, each trial tests solutions on only a single held-out example. We observed that nearly half of the programs that passed this test (45.4\%) failed against the ground-truth rule as judged by the evaluator, inflating acquisition counts without reflecting true rule induction. This limitation highlights that acquisition curves may overstate success when measured on a single example. More robust evaluation schemes that test programs against multiple held-out examples or directly against the underlying rule would better capture whether models achieve genuine inductive generalization.

\section{Implications and Limitations}
\label{discussion}
Our findings demonstrate that hypothesis search achieves human-comparable acquisition curves and outperforms direct program generation on few-shot rule induction. More broadly, this result illustrates how language-based hypothesis generation can serve as a powerful inductive bias for program synthesis, aligning with classic accounts of reasoning as search constrained by structured representations \cite{tenenbaum2011grow}. A distinctive strength of this approach is that input–output examples provide their own validation. Executing a candidate program is sufficient to assess correctness, reducing dependence on external ground-truth annotations and making program induction especially suitable in settings where labeled data are scarce. In addition, program-based hypotheses are inherently interpretable, providing a structured format for inspecting reasoning that is more accessible than end-to-end neural methods.

From an AI perspective, these findings suggest that effective inductive reasoning with LLMs may depend less on end-to-end symbolic inference and more on leveraging language as a flexible inductive bias for program search. Error analysis shows that hypothesis quality is the central bottleneck, and that downstream refinement compensates for errors only at high computational cost. This reflects a broader trade-off in which richer hypothesis representations reduce the burden on search, whereas weaker representations force more extensive search and thereby increase computational overhead, paralleling the classic bias–variance dilemma in learning. Natural language offers a promising substrate for navigating this trade-off, serving as a bridge between the flexibility of statistical learning and the compositional structure of symbolic reasoning. Future progress will depend on improving hypothesis generation and on developing tighter interfaces between natural language and program space to better balance representational richness and search efficiency. In this context, modular pipelines could offer a more scalable alternative to monolithic end-to-end systems, since decomposition into hypothesis generation, compression, and refinement enables targeted improvements to individual components without retraining the entire system.

Another limitation arises from the evaluation protocol, which tests each trial on only a single held-out example. This setup allows incorrect hypotheses to occasionally appear successful, potentially inflating acquisition curves without reflecting true rule induction. More robust protocols with multiple test cases would provide a stronger basis for assessing inductive reasoning in both cognitive models and AI systems, and would better capture whether observed successes reflect genuine generalization rather than overfitting.

Finally, our error analysis relies on an LLM module to evaluate hypotheses against the ground-truth rule descriptions. While this provides a scalable way to assess correctness across thousands of hypotheses, it introduces the possibility of evaluator error. Misjudgments by the evaluator could lead to under- or over-estimating the accuracy of the Generator and Summarizer. Future work could address this limitation by incorporating human evaluation, reducing dependence on a single system’s judgments.

\end{document}